\documentclass[]{fairmeta}
\usepackage{float}
\usepackage{caption}
\usepackage{listings}
\usepackage[ruled,vlined]{algorithm2e}

\title{Matrix: Peer-to-Peer Multi-Agent Synthetic Data Generation Framework}

\author[1,\dagger]{Dong Wang}
\author[1,\dagger]{Yang Li}
\author[1,\dagger]{Ansong Ni}
\author[1]{Ching-Feng Yeh}
\author[1]{Youssef Emad}
\author[1]{Xinjie Lei}
\author[1]{Liam Robbins}
\author[1]{Karthik Padthe}
\author[1]{Hu Xu}
\author[1]{Xian Li}
\author[1]{Asli Celikyilmaz}
\author[1]{Ramya Raghavendra}
\author[1]{Lifei Huang}
\author[1,\dagger]{Carole-Jean Wu}
\author[1,\dagger]{Shang-Wen Li}

\affiliation[1]{FAIR at Meta}

\contribution[\dagger]{Core Contributors}

\abstract{Synthetic data has become increasingly important for training large language models, especially when real data is scarce, expensive, or privacy-sensitive. Many such generation tasks require coordinated multi-agent workflows, where specialized agents collaborate to produce data that is higher quality, more diverse, and structurally richer. However, existing frameworks for multi-agent synthesis often depend on a centralized orchestrator, creating scalability bottlenecks, or are hardcoded for specific domains, limiting flexibility. We present \textbf{Matrix}, a decentralized framework that represents both control and data flow as serialized messages passed through distributed queues. This peer-to-peer design eliminates the central orchestrator. Each task progresses independently through lightweight agents, while compute-intensive operations, such as LLM inference or containerized environments, are handled by distributed services. Built on Ray, Matrix scales to tens of thousands of concurrent agentic workflows and provides a modular, configurable design that enables easy adaptation to a wide range of data generation workflows. We evaluate Matrix across diverse synthesis scenarios, such as multi-agent collaborative dialogue, web-based reasoning data extraction, and tool-use trajectory generation in customer service environments. In all cases, Matrix achieves $2$--$15\times$ higher data generation throughput under identical hardware resources, without compromising output quality.}

\date{\today}
\correspondence{Dong Wang \email{dwoanngg@gmail.com}, Shang-Wen Li \email{shangwel@meta.com}}

\metadata[Code]{\url{https://github.com/facebookresearch/matrix}}
\begin{document}

\maketitle

\section{Introduction}
\label{introduction}

Large scale machine learning models, such as large language models (LLMs) and multi-modal foundation models, are increasingly trained with synthetic data to reduce dependence on costly, noisy, or privacy-sensitive human-curated datasets \cite{grattafiori2024llama3herdmodels, abdin2024phi4technicalreport, openai_dalle3_2023}. 
Recent advances have shifted toward agentic synthetic data generation, where data is produced through interactions among multiple intelligent agents rather than a single model or fixed pipeline. This paradigm enables multi-agent collaboration for diverse generation tasks such as code synthesis, instruction and dialogue creation, knowledge-grounded question answering, and multi-modal content generation. In these settings, 
%LLM-powered agents with complementary roles—such as problem decomposition, code generation, verification, and critique—cooperate to produce higher quality and more verifiable outputs. The resulting 
the workflows often involve complex control flows with loops, moving beyond traditional linear data generation pipelines. For example, Kimi K2~\cite{kimiteam2025kimik2openagentic} employs a large-scale multi-agent data synthesis pipeline to construct diverse tool-use and reasoning demonstrations. Similarly, CWM~\cite{faircodegenteam2025cwmopenweightsllmresearch} leverages autonomous software engineering agents to generate multi-step trajectories for code understanding and debugging.
These systems exemplify the growing adoption of multi-agent pipelines for synthetic data generation in large scale LLM training, underscoring the need for flexible and scalable frameworks for data synthesis.

Generic agent frameworks such as AutoGen~\cite{wu2023autogenenablingnextgenllm, fourney2024magenticonegeneralistmultiagentsolving}, LangGraph~\cite{langgraph2025}, and CrewAI~\cite{crewai2025} provide convenient abstractions for \emph{authoring} agent workflows and expressing control flow. However, operating these workflows at the throughput regime targeted in this work typically requires additional production scaffolding beyond the workflow definition itself, including scalable LLM/tool services, backpressure and concurrency control, retries/timeouts, and logging/metrics. Accordingly, rather than presenting an end-to-end throughput bake-off against these authoring-focused frameworks, we view them as complementary: Matrix can reuse their workflow specifications (e.g., LangGraph-style state graphs) by packaging common patterns as reusable orchestrator subclasses within our runtime, enabling users to design control flow in familiar form while benefiting from Matrix's decentralized scheduling and distributed service integration.

Recently a number of systems have been developed to generate synthetic data for \emph{particular} agentic settings. Notable examples include AgentInstruct~\cite{mitra2024agentinstructgenerativeteachingagentic}, SWE-Agent~\cite{yang2024sweagentagentcomputerinterfacesenable}, SWE-Synth~\cite{pham2025swesynthsynthesizingverifiablebugfix}, TaskCraft~\cite{shi2025taskcraftautomatedgenerationagentic}, and AgentSynth~\cite{xie2025agentsynthscalabletaskgeneration}. These systems demonstrate that carefully designed agent roles and validation loops can produce high-quality data and strong task-level performance. However, their implementations are often optimized around a \emph{specific} task structure (e.g., software engineering, tool execution, or instruction generation), so adapting them to new domains can require non-trivial engineering, such as refactoring the control-flow logic, state representation, and service integration, rather than simply swapping prompts or models.
At scale, running many independent workflow instances also introduces systems challenges. Users typically rely on per-process concurrency (threads/async) and/or external job orchestration (e.g., Kubernetes Jobs, Airflow, or distributed task queues) to execute large numbers of tasks concurrently, and must additionally provision shared services for inference, tool execution and logging.

To address these limitations, we present Matrix, a distributed runtime for scalable, multi-agent synthetic data generation and agentic experimentation. Matrix frames data generation as a \emph{data-to-data transformation}: each input row represents an independent task, and the runtime executes many such tasks concurrently, each running its own agentic workflow.

The core idea behind Matrix is a peer-to-peer (P2P) agent architecture that replaces centralized orchestration with decentralized, message-driven scheduling. The state of each task, which includes orchestration logic, intermediate results, and conversation history, is serialized into messages that are passed among agents. The active agent consumes and updates this state, then emits it to the next agent determined by the orchestrator. Because agents themselves are stateless, they can scale elastically and independently across the cluster.

Unlike traditional batch-level scheduling in distributed execution engines such as Spark~\cite{zaharia2012rdd} and Ray Data~\cite{ray}, where the pipeline controls progress across synchronized batches, Matrix performs row-level scheduling through peer-to-peer message orchestration. Control and data flow are embedded in messages, allowing each task to progress asynchronously through agents. This eliminates idle periods caused by batch-level barriers.

Distributing a centralized workflow with an external job scheduler (e.g., SLURM or Kubernetes) can sidestep a single orchestrator bottleneck by running many workflow replicas in parallel. In practice, however, scaling this way typically pushes substantial distributed-systems work into each application: partitioning the input into jobs, provisioning and load-balancing shared services (LLM inference, tool execution, logging, and state), implementing retries and timeouts, and tuning coupled parameters such as number of jobs, per-job concurrency. Matrix instead provides a self-contained runtime for agentic data generation: workflow state and control are carried as messages, scheduling is decentralized at row granularity, and heavy computation is delegated to shared distributed services with built-in backpressure. This reduces application-specific orchestration glue while scaling to tens of thousands of concurrent workflows.

%Matrix integrates naturally with modern inference engines such as vLLM~\cite{vllm}, SGLang~\cite{zheng2024sglangefficientexecutionstructured}, and leverages Ray~\cite{ray} for distributed execution and containerized environments via Apptainer~\cite{kurtzer2017singularity} for complex services such as software and tools execution.

\textbf{Key Contributions.}
\begin{enumerate}
\item We introduce \textbf{Matrix}, a \textbf{scalable runtime} for large scale multi-agent synthetic data generation capable of efficiently executing tens of thousands of concurrent workflows. Matrix adopts a \textbf{peer-to-peer agent architecture} with message-embedded control and state representation, eliminating centralized orchestration bottlenecks and idle time caused by batch-level synchronization. This design enables fully asynchronous and fine-grained execution at scale.

\item Matrix is designed to be \textbf{flexible} and \textbf{extensible}, supporting diverse multi-agent use cases. Its modular architecture separates key components, including the generation loop and distributed services for LLM inference and containerized execution, and the entire system is fully configurable through Hydra.
%Users can extend the base implementations of the agent and orchestrator to tailor workflow behaviors, and the entire system is fully configurable through Hydra.

\item We evaluate Matrix on three representative case studies: Collaborative Reasoner~\cite{ni2025collaborative}, NaturalReasoning~\cite{yuan2025naturalreasoningreasoningwild28m}, and Tau2-bench~\cite{barres2025tau2}. Matrix achieves \textbf{2--15$\times$ higher token throughput} than specialized baseline systems while maintaining comparable output quality.

\item Matrix is built entirely on an \textbf{open source stack}, including SLURM, Ray, vLLM, SGLang, and Apptainer. It supports both open-weight models and LLM API proxies. We have open sourced the framework to the community to foster open development and collaborative research.
\end{enumerate}

\section{Related Work}

\paragraph{LLM and agentic benchmarks.} 
LLMs are commonly evaluated on reasoning benchmarks 
%that test multi-step logical, mathematical, and professional knowledge capabilities. Examples include 
such as Math-500~\cite{hendrycksmath2021} and MMLU-Pro~\cite{wang2024mmluprorobustchallengingmultitask}. Recent multi-agent systems run on standardized benchmarks that test complex, multi-step reasoning and tool use. Examples include SWE-bench~\cite{jimenez2024swebenchlanguagemodelsresolve}, Tau2-Bench~\cite{barres2025tau2}, MCP-Bench~\cite{wang2025mcpbenchbenchmarkingtoolusingllm}, and MLE-Bench~\cite{chan2025mlebenchevaluatingmachinelearning}. Each benchmark comes with a reference agentic system that solves the tasks, such as SWE-agent and Tau2-agent.
%, and often provides a leaderboard where more sophisticated agents, such as Aria-Dojo~\cite{toledo2025airesearchagentsmachine}, are custom-built to maximize performance. 
%Some benchmarks are also used to generate fine-tuning data, while others, like SWE-Synth~\cite{pham2025swesynthsynthesizingverifiablebugfix}, leverage agents to create additional synthetic data for improving agent performance.
In this work, we use Tau2-Bench and MMLU-Pro as sources of initial tasks to generate agent trajectories that can be used for fine-tuning LLMs.
%In this work, we use Tau2-Bench and MMLU-Pro as sources of initial tasks to generate agent trajectories. These trajectories provide structured demonstrations that can be employed for fine-tuning or for evaluating multi-agent workflows, serving as a practical testbed for Matrix's synthetic data generation capabilities.

\paragraph{Data Synthesis via Multi-agents Workflows.}

The scarcity of high-quality agentic training data has led to the development of synthetic data generation techniques employing multi-agent frameworks. 
%MAG-V~\cite{sengupta2025magvmultiagentframeworksynthetic} is a multi-agent framework that generates synthetic customer-query datasets and reverse-engineers alternative questions for trajectory verification to improve agent performance. 
AgentInstruct~\cite{mitra2024agentinstructgenerativeteachingagentic} generates multi-turn instruction-response data by coordinating multiple agents to propose, verify, and refine synthetic tasks based on seed examples. 
%MATRIX-Gen~\cite{tang2025synthesizingposttrainingdatallms} generates realistic and diverse scenarios with 1000 real-world-grounded agents and structured communication. 
TaskCraft~\cite{shi2025taskcraftautomatedgenerationagentic} automatically generates multi-step, multi-tool agentic tasks with verifiable execution trajectories.%, expanding atomic tasks into structurally and hierarchically complex challenges. 
APIGen-MT~\cite{prabhakar2025apigenmtagenticpipelinemultiturn} is a two-phase framework that generates verifiable, multi-turn agent interaction data.
%by first producing task blueprints with iterative LLM feedback and then simulating human-agent interplay to create complete trajectories.
While these frameworks are tailored to specific data needs and emphasize data quality, our approach offers a generic framework capable of supporting multiple use cases with a focus on scalability.

\paragraph{Peer-to-Peer ML Systems}
Peer-to-peer (P2P) architectures have long been foundational in distributed computing and communications.
%, facilitating decentralized resource sharing and fault tolerance. 
In ML, P2P systems have been leveraged to enhance scalability, privacy, and personalization. For instance, 
%a fully decentralized and asynchronous algorithm~\cite{bellet2018personalizedprivatepeertopeermachine} is introduced for personalized and private ML, employing differential privacy to protect personal datasets while ensuring provable convergence rates. 
The SPIRT~\cite{barrak2023spirtfaulttolerantreliablepeertopeer} framework introduces a serverless P2P ML training architecture that leverages RedisAI for in-database operations, achieving significant reductions in model update times and demonstrating resilience against peer failures. Similarly, 
BlockDFL~\cite{qin2024blockdflblockchainbasedfullydecentralized} employs blockchain-based coordination to facilitate fully decentralized federated learning, incorporating mechanisms to defend against poisoning attacks and reduce communication costs.
%These approaches highlight the potential of P2P systems for addressing challenges such as data privacy and scalability in ML. 
While prior P2P ML systems focus on efficient training and privacy-preserving computation, Matrix introduces a general framework using P2P communication to coordinate agent workflows for scalable multi-agent data synthesis.

\section{Matrix Overview}

This section provides an overview of the Matrix framework, describing its system architecture and the core algorithm that enables scalable, asynchronous multi-agent synthetic data generation.

\subsection{System Architecture}

\begin{figure*}[h]
    \centering
    \includegraphics[width=0.8\textwidth]{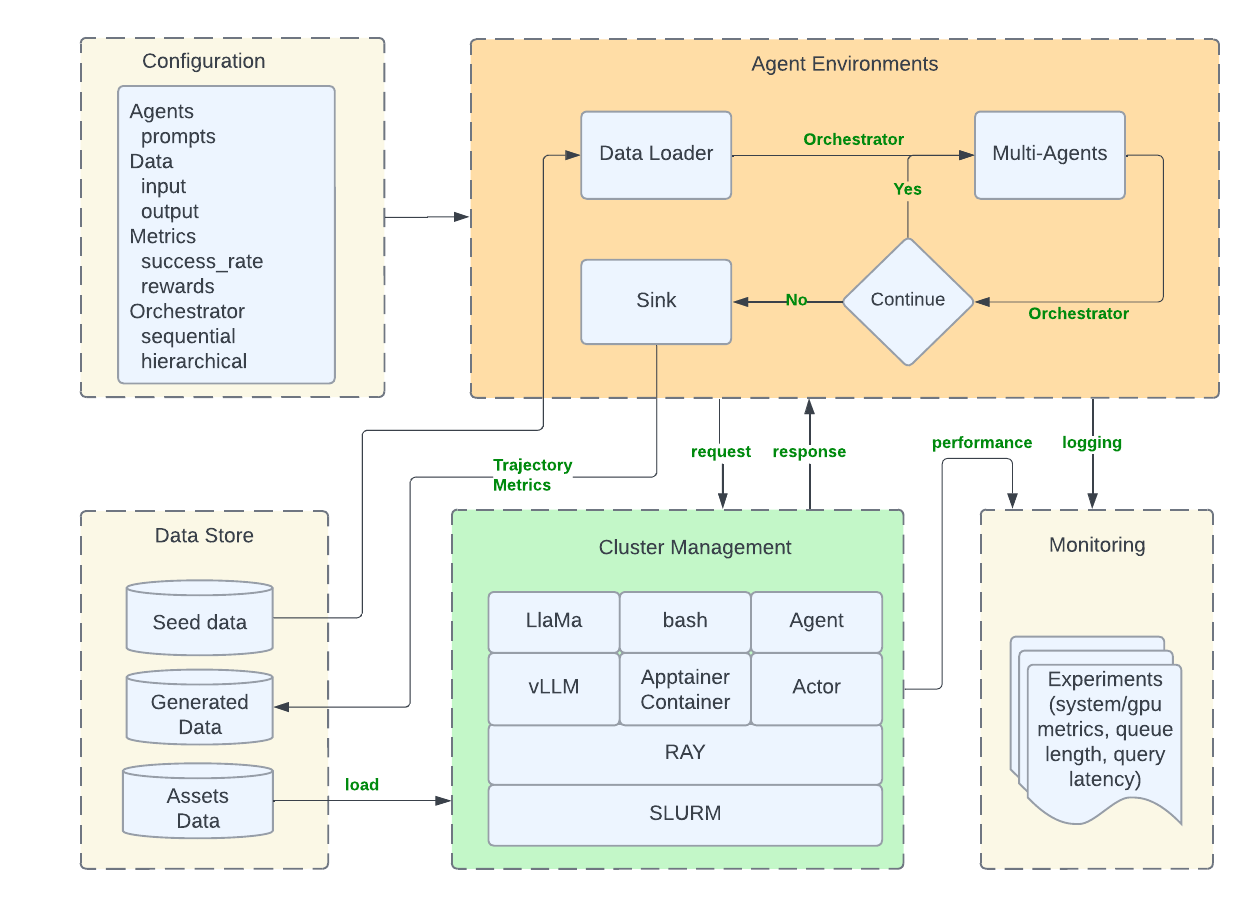}
    \caption{Matrix Agentic Data Generation Architecture.}
    \label{fig:system}
    \vspace{-2mm}
\end{figure*}

Figure~\ref{fig:system} illustrates the architecture of Matrix, a distributed runtime for multi-agent synthetic data generation. The system is designed to be \textbf{modular} and \textbf{configurable}, separating lightweight peer-to-peer agents from scalable backend services (e.g., LLM inference and containerized tool execution) so that components can be adapted and scaled independently across different workflows.

\paragraph{Cluster Management.}
The framework is deployed atop SLURM~\cite{slurm}, a widely adopted distributed computing environment, with a Ray~\cite{ray} cluster serving as the execution substrate. Ray Serve provides high-throughput LLM inference services, backed by vLLM~\cite{vllm}, SGLang~\cite{zheng2024sglangefficientexecutionstructured}, and FastGen~\cite{fastgen2025}. Containerized execution is supported through Apptainer~\cite{kurtzer2017singularity}, enabling stateful environments to be launched on demand. Each agent is implemented as a Ray Actor, allowing scalable parallelization and fine-grained resource placement across worker nodes.

\paragraph{Configuration.}
System configurability is managed through Hydra~\cite{Yadan2019Hydra}, which specifies agent roles, input–output schemas, generation metrics, and resource requirements (e.g., LLM engine selection). The configuration also defines the orchestrator responsible for control and data flow management. Users can define the use case specific orchestrator class in Hydra, then hand-write control flow as imperative if/else logic in the implemetation. Matrix also supports a graph-based specification. For users familiar with LangGraph~\cite{langgraph2025}, we provide a \texttt{LangGraphOrchestrator} that performs state transitions based on a user-supplied LangGraph: users define nodes (agent roles) and edges (transition rules) in the standard LangGraph API, and the orchestrator updates its state by executing the graph’s routing decision at each step. This makes non-sequential workflows easier to express and to visualize while preserving Matrix’s decentralized, message-driven execution model. 

\paragraph{Agents Environments.} 
In the peer-to-peer generation process, each input datum is encapsulated into an orchestrator instance and passed to the initial agent. The agent processes the instance, updates the orchestrator state, and forwards control to the next designated agent. This process continues iteratively until completion. A detailed algorithmic description is provided in Section~\ref{sec:algorithm}. We will use Matrix to build different agents environments in the experiments section. 

\paragraph{Monitoring.} Logging and observability are critical for debugging and performance analysis. Matrix integrates with Grafana~\cite{grafana2025} for real-time monitoring. In addition to standard performance metrics, it provides custom indicators such as distributed queue length and the number of pending asynchronous tasks. These metrics help identify throughput bottlenecks and evaluate overall system health.

% easily parallelize and, enabling flexible scaling of agent behaviors and orchestration across heterogeneous resources. Each agent role can be instantiated as one or more Ray actors, allowing fine-grained control over concurrency and resource allocation. The framework separates configuration, data management, agent environments, and cluster execution into modular components. Each data entry becomes an orchestrator, which governs distributed execution by maintaining and propagating a shared environment state across agents. Agents interact asynchronously through this orchestrator, coordinating tasks and exchanging intermediate results in a scalable manner. LLM inference and containered runtime are built and scaled separately. The system provides monitoring and data logging for both environment-level metrics (e.g., success rates, trajectory statistics) and infrastructure-level metrics (e.g., GPU utilization, queue latency), supporting reproducible and data-driven analysis of multi-agent dynamics.
% Matrix introduces a \textbf{peer-to-peer orchestration model}, where each agent consumes and produces messages representing evolving control and data states.  

\subsection{Data Generation Algorithm}
\label{sec:algorithm}

\begin{algorithm}[h]
\caption{Matrix P2P agentic generation pseudocode.}
\label{alg:code}
\definecolor{codeblue}{rgb}{0.25,0.5,0.5}
\definecolor{gray}{rgb}{0.5,0.5,0.5} % add this in preamble
\lstset{
  backgroundcolor=\color{white},
  basicstyle=\fontsize{7.2pt}{7.2pt}\ttfamily\selectfont,
  columns=fullflexible,
  breaklines=true,
  captionpos=b,
  commentstyle=\fontsize{7.2pt}{7.2pt}\color{codeblue},
  keywordstyle=\fontsize{7.2pt}{7.2pt},
  numbers=left,                  % <-- Add this line
  numbersep=8pt,                 % distance between line numbers and code
  numberstyle=\tiny\color{gray},
}
\centering
\begin{lstlisting}[language=python]
@ray.remote
class AgentActor:  # agent base class with an event loop to process orchestration messages
    async def _event_loop(self, team):
        while True:
            orchestrator = await self.queue.get()
            result = self.process(orchestrator)
            orchestrator.update(result)  # update conversation history and determine next agent
            next_agent = orchestrator.current_agent()
            random.choice(team[next_agent]).send(orchestrator)  # send updated orchestrator to next agent

class SequentialOrchestrator:  # a typical orchestrator with a configurable order of execution
    def update(self, result):
        self.history.append(result)
        self.index = (self.index + 1) % len(self.order)  # take the next agent in the given order with loop around

    def current_agent(self):
        return "_sink" if self.is_done else self.order[self.index]  # loop around until is_done flag is set

def create_team(cfg):  # create a team of agents based on the configuration
    return {
        role: [ray_create_actor(role, role_cfg)  # each agent instance become a Ray actor
               for _ in range(role_cfg.num_instances)]
        for role, role_cfg in cfg.items()
    }
# main processing 
team = create_team(cfg.agents)
for item in dataset:  # process each dataum concurrently up to max_concurrency asyncio tasks
    orchestrator = Orchestrator(item)
    first = random.choice(team[orchestrator.current_agent])
    first.send(orchestrator)  # send the orchestrator to the start agent
\end{lstlisting}
\end{algorithm}

Algorithm~\ref{alg:code} illustrates the core workflow of Matrix’s peer-to-peer agentic generation runtime. The system begins by reading the Hydra configuration \texttt{cfg}, which specifies all agent roles and their resource requirements (e.g., CPU, GPU, and memory). As shown in Lines 19–24, the function \texttt{create\_team()} instantiates a distributed team of Ray actors for each agent role, allowing heterogeneous resource allocation across agent types.

\paragraph{Agent EventLoop.} The main generation loop (Lines 26–30) iterates over input items in the dataset. For each item, an \texttt{Orchestrator} object is created to manage task-specific state and control flow. The orchestrator is initially dispatched to the first agent in the sequence, sampled randomly from the corresponding role group. Each agent runs as a persistent event-driven process (Lines 3–9) implemented by the \texttt{AgentActor} class. Within its asynchronous \texttt{\_event\_loop}, the agent dequeues orchestrators from its inbox, applies role-specific logic through \texttt{process()}, updates task state, and forwards it to the next designated agent.

\paragraph{Orchestration.} An example orchestrator \texttt{SequentialOrchestrator} is in Lines 11–17. It maintains a structured \texttt{history} of intermediate results and a configurable \texttt{order}, which determines the sequence of participating agents. After each interaction, \texttt{update()} advances the internal index to the next agent. The process continues cyclically until the orchestrator’s \texttt{is\_done} flag is set, at which point it is routed to a special terminal agent, \texttt{\_sink}, for result persistence and metric aggregation.

\paragraph{Concurrency Control.} Advanced runtime features (not shown for brevity) include task-level concurrency control through a \texttt{max\_concurrency} parameter and semaphore-based scheduling. The semaphore is decremented when an orchestrator is dispatched to the \texttt{first} agent and incremented upon completion by the \texttt{\_sink} agent. This mechanism limits the number of active orchestrators, ensuring controlled resource utilization and stability during large-scale distributed execution. Note we rely on Ray to avoid race conditions in distributed RPC calls.

\section{Agent Environment Design for Matrix}

This section describes the system’s internal design, including its orchestration model, distributed service layer, parallelism strategies, scheduling policies, fault tolerance mechanisms and network bandwidth optimization.

\subsection{P2P Orchestration}
\label{sec:orchestration}

\begin{figure*}[h]
     \centering
     \begin{subfigure}[b]{0.48\textwidth}
         \centering
         \includegraphics[width=0.8\textwidth]{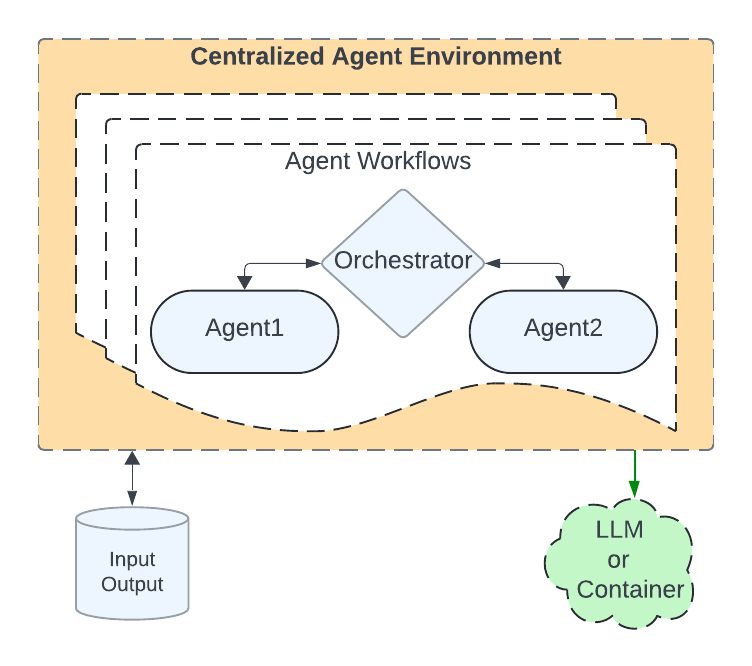}
    \caption{Traditional centralized orchestration.}
    \label{fig:centralized}
     \end{subfigure}
     \hfill
     \begin{subfigure}[b]{0.48\textwidth}
         \centering
    \includegraphics[width=0.8\textwidth]{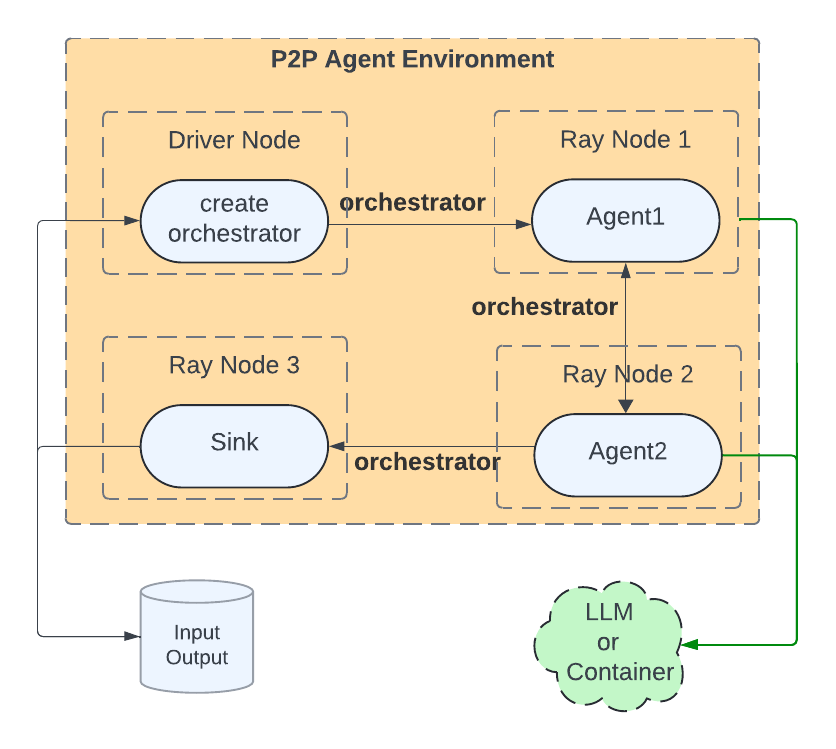}
    \caption{P2P Orchestration in Matrix.}
    \label{fig:p2p_vs_centralized}
     \end{subfigure}
     \caption{Compare Centralized vs P2P Orchestration.
     }
     \label{fig:plots}
\end{figure*}

As illustrated in Figure~\ref{fig:centralized}, centralized orchestration must manage execution order (control flow), message passing (data flow), and the full lifecycle of requests and responses for LLMs and containerized environments. Handling all of this for tens of thousands of concurrent workflows quickly becomes a scalability bottleneck. Matrix addresses this by representing workflows as serializable \emph{orchestrators} that can be updated and exchanged among distributed agents (Figure~\ref{fig:p2p_vs_centralized}). The \emph{driver}, which runs the generation framework, plays a lightweight role: it simply publishes an orchestrator to start a task, enabling an asynchronous initiation model. Agents equipped with LLMs and tools consume messages, perform local actions, update both control and data states, and forward the updated orchestrator to the next agent. Execution continues until the orchestrator signals completion, at which point a designated sink collects the final message and persists it to the output dataset. Using P2P orchestration, Matrix avoids bottlenecks, improves scalability, and enables fully asynchronous execution among agents.

\subsection{Distributed Services}

Matrix offloads computationally intensive tasks to distributed services, allowing them to scale independently of the agents. For LLM inference, Matrix employs gRPC-based communication to avoid HTTP overhead. Because the Ray head node can become network-bound, Matrix maintains a local cache of active model replica URLs, enabling direct load-balanced traffic through worker nodes. Sticky routing can reuse prefix cache for multi-turn long conversations. In addition to Huggingface models, proxies are built for commercial LLM API services.
For stateful services such as Apptainer containers, agents acquire containers by ID to be able to route multiple commands to the same container instance, rather than a randomly selected one. This is managed via a \emph{resource pool} and a \emph{registry} that maps container IDs to Ray actors running the corresponding containers. This design allows agents to efficiently route messages and reuse shared resources.

\subsection{Parallel Execution Strategies}
\label{sec:parallelism}

Matrix supports multiple forms of parallelism to maximize scalability and cluster utilization.

\begin{itemize}
    \item \textbf{Data parallelism.} Similar to distributed processing systems such as Spark~\cite{zaharia2012rdd} and Ray Data~\cite{ray}, Matrix can partition large input datasets consisting of many small files for independent processing. For multi-file inputs, Matrix automatically distributes files across partitions. Datasets containing a few large files can be preprocessed into smaller shards to enable higher parallelism.  

    \item \textbf{Task parallelism.} Multiple generation tasks can execute concurrently using asynchronous programming, threads, or processes. Matrix adopts an \texttt{asyncio}-based model: the driver initializes orchestrators, and agents process tasks asynchronously. Since computationally heavy operations are offloaded to distributed services, lightweight agents can handle tens of thousands of concurrent tasks efficiently without I/O blocking.  

    \item \textbf{Agent parallelism.} Each agent role is implemented as Ray actors with configurable CPU, GPU, and memory allocations. Roles can scale horizontally by launching multiple distributed agent instances, each processing assigned tasks independently. Ray system distributes these actors across cluster nodes, enabling each role to scale without the resource contention commonly seen in centralized orchestration.
    
\end{itemize}

For LLM-based agents, computational cost dominates over input pipeline overhead. Usually data loading is not a bottleneck (one exception is the NaturalReasoning task in Section~\ref{sec:nr}). Matrix’s peer-to-peer architecture and distributed services ensure efficient utilization of cluster resources even with moderate data and agent-level parallelism. This efficiency arises from Matrix’s ability to run tens of thousands of asynchronous tasks concurrently, each processing one data item independently.

\subsection{Row-Level Scheduling}
\label{sec:row_vs_batch}

In batch processing systems, such as Ray Data, tasks are grouped into fixed-size batches and executed by actors. While this approach can reduce per-task scheduling overhead for homogeneous workloads, it introduces inefficiencies when tasks have variable computational demands or diverging control flows. A long-running or complex task within a batch can keep the current batch running and stall the execution of subsequent batches, creating idle resources and underutilized GPUs. We refer to this phenomenon as \textit{batch-level scheduling}. 

In contrast, Matrix schedules each task independently as soon as prior tasks complete, a mechanism called \textit{row-level scheduling}. Each orchestrator message representing a single task flows through the P2P agent network. This design eliminates the bubble effects inherent in batch processing, achieves higher GPU utilization, and reduces end-to-end latency for heterogeneous, multi-agent workloads. Row-level pipelining, combined with distributed services and asynchronous agent execution, is a key factor in Matrix's scalability and efficiency for large-scale data synthesis tasks.

\subsection{Agent Fault Tolerance}
\label{sec:fault}

Matrix currently provides \emph{at-most-once} execution semantics. Tasks may fail for various reasons, including network errors, timeouts, and actor crashes. Failed tasks can be collected from the output dataset and re-run offline if needed.
Matrix workflows are implemented by extending a base agent class, and use-case-specific logic may introduce bugs that crash an agent. Ray can restart crashed agent actors, however, any in-flight orchestrator messages that were dequeued by the crashed agent are not recoverable under at-most-once semantics.
To track in-flight orchestrators and surface failures reliably, Matrix uses per-role message brokers. All agents of the same role share a broker, and all incoming and outgoing orchestrator messages for that role are routed through it. Each broker maintains (i) an incoming queue of orchestrators waiting to be processed and (ii) an assignment map that records which orchestrators are currently assigned to which agent instance. The broker dispatches orchestrators to agents in a round-robin manner. After an agent finishes processing an orchestrator, it returns the updated orchestrator to the broker, the broker then removes the corresponding entry from the assignment map and forwards the orchestrator to the next role's broker.
When an agent crashes and is restarted by Ray, it re-registers with its broker. The broker detects that the previous instance has died, marks all orchestrators assigned to that instance as failed based on the assignment map, and forwards them to the sink for persistence as failed trajectories. With this design, use-case-specific agents can crash and restart without halting the system, as long as the brokers and sink remain available.
Brokers and the sink are framework components (not customized per use case), and we rely on them for reliability. If a broker or the sink fails, the generation job terminates. To mitigate transient network issues, Matrix uses retries for communication between agents, brokers and sink.

\subsection{Message Offloading}
\label{sec:message_offloading}
The orchestrator is serialized and exchanged among agents. As shown in Algorithm~\ref{alg:code}, its \texttt{history} field stores inter-agent conversations, which can be large. A common optimization is to offload this history to an external cache such as Redis. While this reduces orchestrator size, it simply shifts network traffic from occurring between agents to occurring between agents and the cache. Since the history is frequently updated and used for constructing LLM prompts, the total network bandwidth can actually double because each agent must retrieve, update, and store the complete history every turn.

Matrix instead retains the history structure within the orchestrator, while storing large conversation content that exceed a configurable size threshold in Ray’s distributed object store. The history holds only the object identifiers, and content is retrieved on demand. Objects are immutable once stored, and all history-related objects are deleted when the orchestrator signals completion. This design keeps the orchestrator compact, reduces redundant transfers, and minimizes network load. Section~\ref{sec:tau2_size} quantifies these benefits experimentally.

\subsection{System Debugging}
\label{sec:debugging}

Debugging distributed systems is challenging, especially under peer-to-peer message passing. Matrix relies on structured logging and trajectory recording for debuggability. Ray streams actor logs back to the driver process, enabling a ``local-like'' debugging experience even when agents are distributed across the cluster.
Matrix also records a full trajectory for each input task. When a task encounters an issue, the trajectory includes the relevant error context for offline analysis (e.g., timeouts, connection failures, and service errors).
For unexpected exceptions, including agent implementation bugs, each agent runs an \texttt{asyncio} event loop and tracks pending futures. Unhandled exceptions propagate to the corresponding future. Matrix then marks the orchestrator as failed and routes it to the sink, which persists the failed trajectory to the output dataset. Users can subsequently filter failed trajectories and re-run them if needed.

\section{Experiments}

We evaluate Matrix across three case studies on synthetic data generation. Together, these experiments demonstrate the framework’s scalability, robustness, and adaptability to diverse workloads. In this section, the terms “Matrix” and “P2P-agent” are used interchangeably to refer to the same framework.

\subsection{Collaborative Reasoner (Coral)}

\begin{figure*}[h]
     \centering
     \begin{minipage}[t]{0.48\textwidth}
         \centering
         \includegraphics[width=0.9\textwidth]{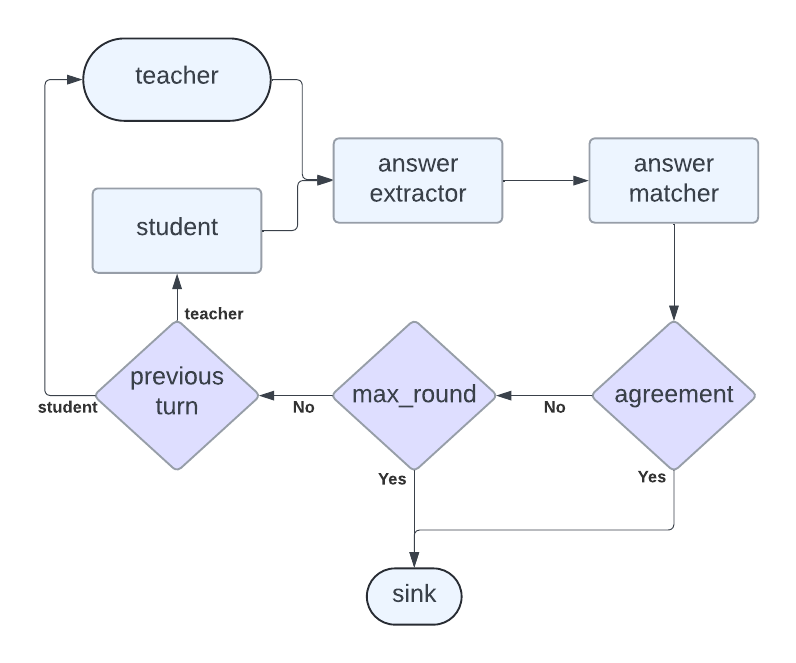}
    \caption{P2P-agents for Collaborative Reasoner.}
    \label{fig:coral_arch}
     \end{minipage}
     \hfill
     \begin{minipage}[t]{0.48\textwidth}
         \centering
    \includegraphics[width=0.9\textwidth]{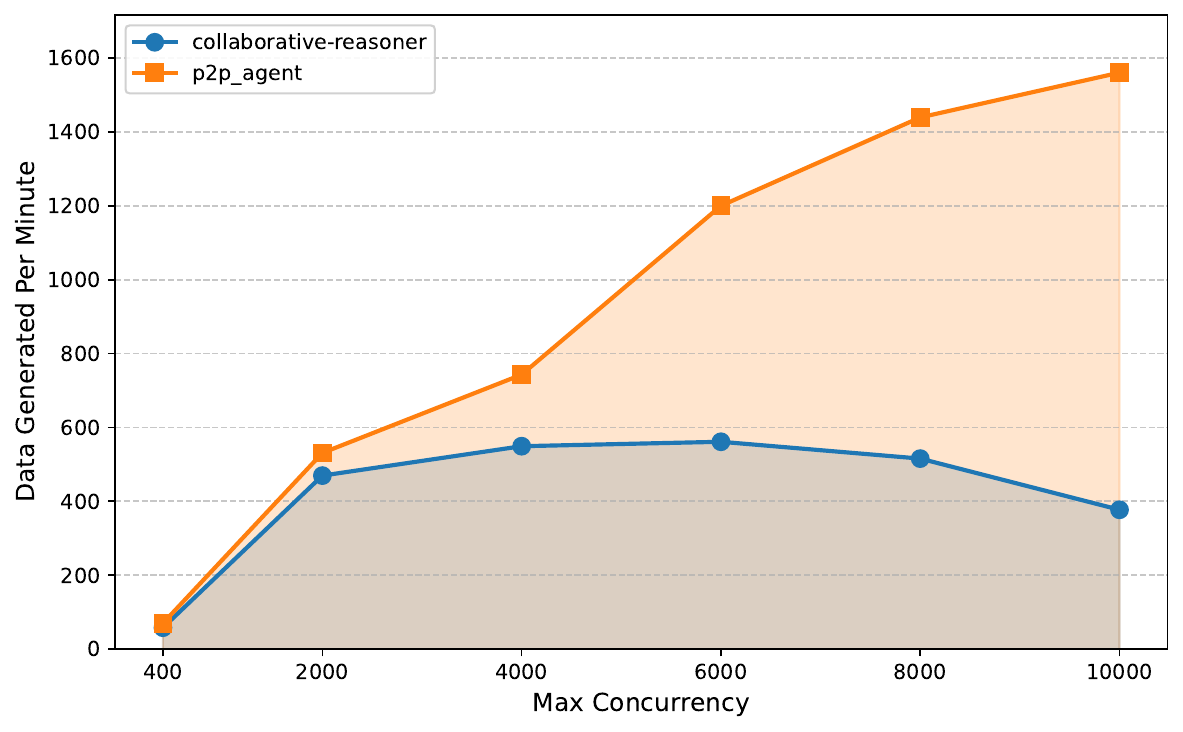}
    \caption{Scalability of P2P-agents vs Coral baseline.}
    \label{fig:coral}
     \end{minipage}
\end{figure*}

Collaborative Reasoner (Coral)~\cite{ni2025collaborative} evaluates and improves multi-agent collaborative reasoning in LLMs through dialogue-driven tasks. Unlike single-agent evaluations, Coral requires two agents to discuss, disagree, and reach consensus over multi-turn interactions. Scalable training data is generated via self-collaboration, where an LLM plays both roles. In this work, we adopt the same agent setup, implemented as distributed agents in Figure~\ref{fig:coral_arch}.

We directly compare Matrix to the official Collaborative Reasoner implementation~\cite{ni2025collaborative_github} as the baseline. Both systems use asyncio for concurrency. The baseline framework uses a single orchestrator to coordinate thousands of concurrent generation tasks, while Matrix distributes coordination responsibilities across agents in a peer-to-peer fashion. To compare the two, we run the same number of MMLU-Pro questions by changing the number of A100 nodes, and in both cases use Llama-3.1-8B-Instruct~\cite{grattafiori2024llama3herdmodels} as the underlying language model for all agents. Task concurrency is adjusted according to the number of A100 nodes as $50 \times N_{GPU}$, leveraging all 8 GPUs per node with 50 concurrent queries per GPU.
As shown in Figure~\ref{fig:coral}, the Matrix implementation scales almost linearly as more GPU nodes are added, while the centralized orchestration approach of the baseline system becomes a bottleneck and plateaus due to the overhead of scheduling a large number of asynchronous tasks from a single control point.

 \paragraph{Large-Scale Results.}
We further tested both systems on 31 A100 nodes (248 GPUs) using LLaMA-3.1-8B-Instruct. For P2P-agent, we set the concurrency to $248 \times 50 \equiv 12,400$, while Coral was configured with its optimal concurrency of 5,000 based on Figure~\ref{fig:coral}. As shown in Table~\ref{tab:coral_1m}, P2P-agent generates 2B tokens in 4 hours, achieving $6.8\times$ higher throughput than the official Coral implementation on the same hardware. Importantly, both systems attain nearly identical agreement correctness, the metric used to measure data quality, consistent with Coral’s reported result of 0.456 for LLaMA-3.1-8B-Instruct~\cite{ni2025collaborative}.
\begin{table}[h]
\centering
\caption{P2P-Agent achieves \textbf{6.8}$\times$ higher token throughput than Coral baseline.}
\begin{tabular}{l r r}
\toprule
\textbf{Metric} & \textbf{Coral Baseline} & \textbf{P2P-Agent} \\
\midrule
Runtime & 9:03:22 & 4:17:05 \\
Concurrent tasks & 5{,}000 & 12{,}400 \\
Total trajectories & 300k & 1 Million \\
Agreement correctness & 0.4732 & 0.4778 \\
Tokens generated & 616{,}759{,}036 & 2{,}002{,}025{,}810 \\
Tokens per second & 18{,}917 & 129{,}833 \\
\bottomrule
\end{tabular}
\label{tab:coral_1m}
\end{table}

\subsubsection{Overhead Analysis}

We analyze system performance to identify overhead and potential bottlenecks in Matrix. Unless otherwise noted, experiments use 8 H100 nodes (64 GPUs) to generate 200k Coral trajectories.

\paragraph{Latency breakdown.}
We find that Matrix incurs minimal queuing and orchestration overhead at scale. We instrument end-to-end task latency and attribute it to: (i) agent processing, (ii) queuing delay, and (iii) task initialization. Table~\ref{tab:latency} reports the breakdown over all trajectories, and Table~\ref{tab:latency_p90} reports the breakdown for the slowest 10\% of trajectories. For typical trajectories, agent processing accounts for $\sim$80\% of end-to-end latency and queuing is negligible. For the slowest trajectories, processing dominates even more ($\sim$99\%).
\begin{table}[h]
\centering
\begin{minipage}[t]{0.48\textwidth}
\centering
\caption{Latency breakdown.}
\begin{tabular}{lrrr}
\hline
\textbf{Stage} & \textbf{Median} & \textbf{P90} & \textbf{P99} \\
\hline
Agent processing & 80.12\% & 99.30\% & 99.92\% \\
Queuing & 0.0289\% & 0.851\% & 5.73\% \\
Initialization & 0.0051\% & 1.18\% & 7.34\% \\
\hline
\end{tabular}
\label{tab:latency}
\end{minipage}
\hfill
\begin{minipage}[t]{0.48\textwidth}
\centering
\caption{Latency breakdown for slow tasks.}
\begin{tabular}{lrrr}
\hline
\textbf{Stage} & \textbf{Median} & \textbf{P90} & \textbf{P99} \\
\hline
Agent processing & 99.72\% & 99.92\% & 99.97\% \\
Queuing & 0.00172\% & 0.025\% & 0.093\% \\
Initialization & 0.000005\% & 0.034\% & 0.128\% \\
\hline
\end{tabular}
\label{tab:latency_p90}
\end{minipage}
\end{table}

\paragraph{Network bandwidth}
We estimate the network bandwidth required to transmit orchestration messages. Under the Coral workload, peer-to-peer orchestration generates 2.26M serialized messages and consumes $\sim$1.6~MB/s of network bandwidth (median $\sim$1.63~MB/s; P99 $\sim$3.47~MB/s), indicating that orchestration traffic is modest relative to cluster network capacity.

\paragraph{Bottleneck study}
To isolate Matrix runtime overhead from model inference cost, we construct dummy Coral agents that do not invoke an LLM and instead return pre-formatted text by concatenation. We ensure that the synthetic responses match the expected response lengths and turn structure, yielding a ``best-case compute'' setting that exposes runtime bottlenecks. In this configuration, the system sustains $\sim$1.1k trajectories/s and processes \textbf{12k} orchestration messages per second. The corresponding estimated network bandwidth for orchestration is $\sim$77.9~MB/s (median $\sim$82.9~MB/s; P99 $\sim$97~MB/s). As shown in Table~\ref{tab:dummy_agent}, agent processing drops to $\sim$37\% of end-to-end latency, task initialization becomes visible, and queuing remains small. The remaining overhead likely comes from RPC, serialization, and network costs. While this experiment suggests a limit of roughly 12k orchestration messages/s per run, Matrix can exceed this throughput via data parallelism, as discussed in Section~\ref{sec:nr}.
\begin{table}[h]
\centering
\caption{Latency breakdown of dummy agents without real compute.}
\begin{tabular}{lrrr}
\hline
\textbf{Stage} & \textbf{Median} & \textbf{P90} & \textbf{P99} \\
\hline
Agent processing & 36.72\% & 62.81\% & 82.63\% \\
Queuing & 0.074\% & 1.13\% & 6.95\% \\
Initialization & 0.768\% & 10.72\% & 24.51\% \\
\hline
\end{tabular}
\label{tab:dummy_agent}
\end{table}

\subsubsection{Actor Crash Recovery}

We evaluate robustness by generating 200k Coral trajectories under two settings: (i) no faults, and (ii) injected faults where we randomly kill an agent actor every 12 minutes. Under the at-most-once semantics described in Section~\ref{sec:fault}, killing an actor may drop any in-flight orchestrators assigned to it, we therefore report the number of lost tasks. As shown in Table~\ref{tab:kill}, actors are killed 7 times and each time Ray restarts them within seconds on average. Table~\ref{tab:recovery} shows that approximately 2\% of tasks are lost in the fault-injection setting, while throughput decreases by only 5\%.
\begin{table}[h]
\centering
\centering
\begin{minipage}[t]{0.48\textwidth}
\caption{Coral actors restarts.}
\begin{tabular}{lrrr}
\hline
\textbf{Agent} & \textbf{Restarts} & \textbf{Duration} & \textbf{Lost Tasks} \\
\hline
answer & 2 & 0.322 & 424 \\
~~\_extractor & & & \\
answer & 2 & 0.000 & 2 \\
~~\_matcher & & & \\
student          & 1 & 2.304 & 1474 \\
teacher          & 2 & 2.069 & 2180 \\
\hline
\end{tabular}
\label{tab:kill}
\end{minipage}
\hfill
\begin{minipage}[t]{0.48\textwidth}
\centering
\caption{Impact of agent restarts.}
\begin{tabular}{l r r}
\toprule
\textbf{Metric} & \textbf{No Crash} & \textbf{With Crash} \\
\midrule
Runtime & 1:26:16 & 1:18:43 \\
Total trajectories & 200k & 200k \\
Lost trajectories & 0 & 4080 \\
Agreement correctness & 0.4781 & 0.4856 \\
Tokens generated & 391{,}200{,}916 & 340{,}338{,}986 \\
Tokens per second & 75{,}579 & 72{,}059 \\
\bottomrule
\end{tabular}
\label{tab:recovery}
\end{minipage}
\end{table}

\subsection{NaturalReasoning}
\label{sec:nr}

NaturalReasoning~\cite{yuan2025naturalreasoningreasoningwild28m} is a large-scale dataset designed to advance the reasoning capabilities of LLMs across diverse domains, including STEM, Economics, and Social Sciences. It contains 2.8M challenging questions generated automatically by LLMs. These questions are extracted and synthesized from pretraining corpora, ensuring high diversity and difficulty. Models fine-tuned on NaturalReasoning demonstrate improved sample efficiency and reasoning accuracy compared to prior datasets.
In this experiment, we use Matrix to curate a NaturalReasoning-style dataset from raw web documents. This workflow stresses Matrix in a different regime than multi-turn dialogue: most inputs are filtered out early, while the remaining fraction triggers expensive downstream processing. The curation pipeline consists of three agents, as illustrated in Figure~\ref{fig:nr}:
\begin{itemize}
    \item \textbf{Filter:} English-language web documents are identified, and a fine-tuned LLaMA-3.1-3B-Instruct model classifies whether a document contains reasoning content. The classifier is trained on a subset of NaturalReasoning examples as positives and randomly sampled web documents as negatives.
    \item \textbf{Score:} Each document is evaluated along multiple quality axes using LLaMA-3.3-70B-Instruct, following prompts derived from the original NaturalReasoning methodology.
    \item \textbf{Question:} Questions are extracted from the filtered web documents, reference answers are identified when available, and independent reasoning steps leading to a final answer are generated, all using LLaMA-3.3-70B-Instruct. Optionally, we grade the extracted answer and check its consistency against the independently generated answer to further filter low-quality examples.
\end{itemize}
 
\begin{minipage}[b]{0.48\textwidth}
    \centering
    \includegraphics[width=0.9\linewidth]{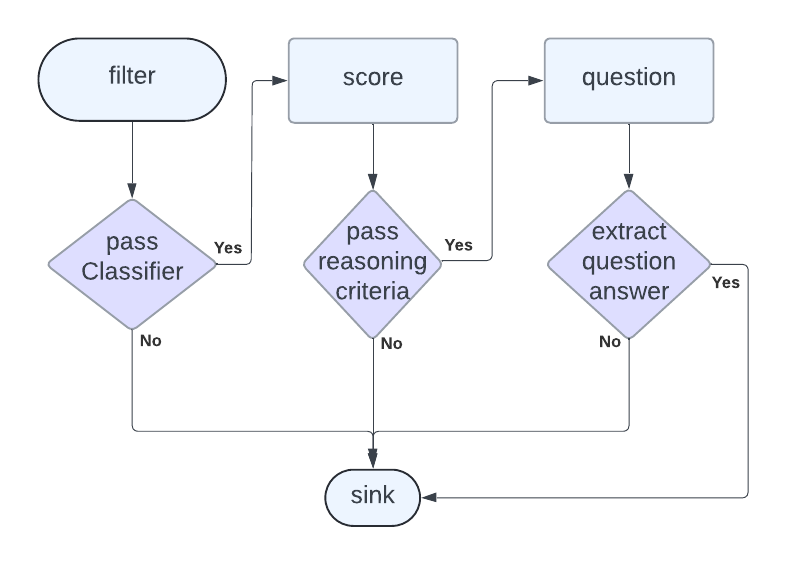}
    \captionof{figure}{P2P-agents for NaturalReasoning data curation.}
    \label{fig:nr}
\end{minipage}
\hfill
\begin{minipage}[b]{0.48\textwidth}
    \centering
    \begin{tabular}{l r}
    \toprule
    \textbf{Filter step} & \textbf{Percentage} \\
    \midrule
    filter\_by\_en & 3.68 \\
    filter\_by\_classifier & 90.24 \\
    filter\_by\_score & 0.44 \\
    filter\_by\_no\_boxed\_answer & 0.19 \\
    success & 5.45 \\
    \bottomrule
    \end{tabular}
    \captionof{table}{Filtering statistics on 25M DCLM web documents.}
    \label{tab:nr-filter}
\end{minipage}

For large-scale curation, we process up to 25M web documents from DCLM~\cite{li2025datacomplmsearchgenerationtraining}. The 3B filter model is efficient because most documents are rejected with a single-token (Yes/No) output. Overall, 5.45\% of documents pass all filters, yielding approximately 1M high-quality reasoning questions and answers (Table~\ref{tab:nr-filter}).

\subsubsection{Evaluating Parallelism and Throughput}
\label{sec:nr-ablation}
Using a 500k DCLM subset, we evaluate the impact of the three types of parallelism supported by Matrix represented as a tuple (data parallelism, task parallelism, and agent parallelism) in Table~\ref{tab:nr-ablation}. We deployed 32 A100 nodes with 8 GPUs each. The fine-tuned 3B model was replicated 32 times, while the 70B model used 56 replicas. We set the maximum concurrent tasks to be 14k. The estimated concurrent requests per 70B replica is
$14k \times (1-3.68\%-90.24\%) \div 56 \approx 15$, which can maintain high GPU utilization without introducing long latencies or timeouts. The 3B model in Filter agents are not the bottleneck even though they handle $97\%$ of the data after English filter.
\begin{table}[h]
\centering
\caption{P2P-agent throughput for 500k webdoc.}
\begin{tabular}{l c r}
\toprule
\textbf{Settings Name} & \textbf{Three Parallelisms}  & \textbf{Normalized Throughput} \\
\midrule
1 & (1, 14000,1) & 1 \\
2 & (20, 700, 1) & 1.61 \\
\midrule
3 & (240, 1, 1) & 0.38 \\
4 & (240, 50, 1) & 1.43 \\
\midrule
5 & (1, 14000, 2) & 1.03 \\
6 & (1, 14000, 10) & 0.91 \\ 
\bottomrule
\end{tabular}
\label{tab:nr-ablation}
\end{table}

\textbf{Data parallelism.} The first two settings present the results for data parallelism. In Setting 1, although the system was configured to allow up to 14k concurrent tasks, only about 700 were observed during the experiment, which is well below the target concurrency. This shortfall occurs because 93\% of the input documents are filtered out early (Table~\ref{tab:nr-filter}), so that the input pipeline can not keep up with the Filter agent. To address the input bottleneck, we increased data parallelism by splitting the dataset into 20 partitions for Setting 2. This raises the effective concurrency to $20 \times 700 \equiv 14k$, matching our target. This adjustment yields a $1.61\times$ speedup, demonstrating how data parallelism helps alleviate the input pipeline bottleneck. Increasing the number of partitions beyond 20 provides little additional benefit, since task-level parallelism within each partition already saturates the GPUs.

\textbf{Task parallelism.} Comparing Settings 3 and 4, running 50 concurrent tasks per data partition yields a 3.8$\times$ speedup compared to single-task execution, even with 240 data partitions. This result shows that increasing asynchronous task concurrency is more effective than simply creating a larger number of data partitions. Moreover, further increasing data parallelism would require additional agent instances, which in turn demands more CPU resources.

\textbf{Agent parallelism.} Comparing Settings 1 and 5, doubling the number of agent instances (excluding the sink) results in a modest throughput gain; while Setting 6 shows further increasing agent instances has no benefits. This is because LLM inference is handled by Ray Serve, agents remain I/O-bound. While increasing the number of instances offers limited benefit for the NaturalReasoning workflow, Matrix can efficiently scale agent instances when agents perform heavier CPU or GPU computations, highlighting the framework’s flexibility and readiness for diverse workloads.

Although the design space of the three kinds of parallelism can be huge, our setup prefers 14k max concurrency given the number of GPUs. We further determined 700 as the maximum achievable asyncio task concurrency per data partition. Moreover, increasing data partitions beyond 20 or increasing agent parallelism beyond 2 has small effect on throughput. Because of the peer-to-peer architecture, task parallelism alone often achieves high resource utilization. Therefore, small degrees of data and agent parallelism are typically sufficient as the initial configuration for new use cases.

\subsubsection{Impact of Scheduling Granularity}
\label{sec:nr-scheduling}

We compare Matrix's \emph{row-level} scheduling to a \emph{batch-level} baseline implemented with Ray Data (Algorithm~\ref{alg:baseline_code}). We emphasize that Ray Data is a general-purpose batch processing engine designed primarily for data-parallel ETL and batched model inference. Our goal in this comparison is \emph{not} to claim that Ray Data is an optimized framework for agentic workflows, but to use it as a representative and widely used batch-oriented alternative for practitioners building scalable LLM-calling pipelines on Ray.

In the Ray Data baseline, each batch is processed by a Ray actor \texttt{BatchProcessing} (Lines 1--12), which launches multiple asynchronous tasks to process rows concurrently (Line 8). Each task executes an agentic workflow (Lines 10--12) that is functionally similar to the P2P-agent logic, except that (i) all agents are co-located within the same actor process and (ii) orchestration is implemented within the batch processor rather than being carried by peer-to-peer messages.

This baseline removes the single centralized orchestrator bottleneck and distributes orchestration across many CPU workers, each responsible for one batch. However, because multi-agent workflows have \emph{data-dependent control flow} (e.g., branching, retries, early termination, and variable numbers of steps), conventional batch-inference optimizations are difficult to apply: different rows within a batch may invoke different agents and different numbers of LLM/tool calls. As a result, even under Ray Data, each row must effectively be executed as an independent asynchronous workflow, and the batch mainly serves as a scheduling container rather than enabling true batched execution of LLM calls.
% Consequently, the primary distinction in this experiment is the \emph{scheduling granularity}: Ray Data introduces batch-level barriers in which a new batch is scheduled only after all rows in the current batch finish, while Matrix schedules each row independently and can immediately admit new work as soon as any row completes. This experiment is therefore intended to isolate the impact of batch-level synchronization on heterogeneous, multi-step LLM-calling workloads, rather than to evaluate Ray Data for its intended batched inference use cases.

\begin{algorithm}[h]
\caption{Pseudo-code of Ray Data Baseline.}
\label{alg:baseline_code}
\definecolor{codeblue}{rgb}{0.25,0.5,0.5}
\definecolor{gray}{rgb}{0.5,0.5,0.5} % add this in preamble
\lstset{
  backgroundcolor=\color{white},
  basicstyle=\fontsize{7.2pt}{7.2pt}\ttfamily\selectfont,
  columns=fullflexible,
  breaklines=true,
  captionpos=b,
  commentstyle=\fontsize{7.2pt}{7.2pt}\color{codeblue},
  keywordstyle=\fontsize{7.2pt}{7.2pt},
  numbers=left,                  % <-- Add this line
  numbersep=8pt,                 % distance between line numbers and code
  numberstyle=\tiny\color{gray},
}
\centering
\begin{lstlisting}[language=python]
@ray.remote
class BatchProcessing:  # base class to run as a Ray actor
    def __call__(self, batch):
        async def _process_batch(rows):
            tasks = [self.process(row) for row in rows]
            return await asyncio.gather(*tasks)  # use asyncio to process all tasks in the batch

        return asyncio.run(_process_batch(batch))

    async def process(self, row: Dict[str, Any]):  # base class method to be overwritten for each use case
        """abstract method to process one input task"""
        pass
    
ds = ray.data.read_json(data_dir)  # read input jsonl files into Ray data
output = ds.map_batches(  # split input to batches for concucurrent processing
    BatchProcessing,
    batch_size=cfg.batch_size,
    num_cpus=1,
    concurrency=cfg.data_parallelism  # max number of batches to run concurrently
)
\end{lstlisting}
\end{algorithm}

\paragraph{Large-Scale Results.}

We then compare Matrix P2P-agent with the Ray Data baseline to run large scale curation over DCLM up to 25M web documents. Both setups utilize the same GPU resources and 14k concurrent tasks. For the P2P-agent configuration, we adopt Setting 2, i.e., (20, 700, 1), from Table~\ref{tab:nr-ablation}. For the Ray Data baseline, we use Setting 4, i.e., (240, 50, 1). Through experiment, Setting 2 with 700 as batch size would result in peaks and valleys in GPU requests, the smaller batch size of 50 in Setting 4 can smooth GPU requests. The two setups have similar throughputs in P2P-agent experiment and the latter fits Ray Data based implementation.

Each setup is executed for over 10 hours, measuring token throughput. Results in Table~\ref{tab:nr_1m} show that P2P-agent achieves 2.1$\times$ higher token throughput than the batch-level baseline. The efficiency gap stems from scheduling granularity: in batch-level scheduling, a new batch cannot begin until all tasks in the current batch complete. Due to control divergence and variable task length, a few slow tasks in a batch block downstream processing, creating idle GPU time. In contrast, row-level scheduling in P2P-agent allows each completed row to immediately trigger the next task without waiting for others, fully utilizing compute resources. Similar behaviour has been observed in LLM inference systems, where “continuous batching” or token-level scheduling can replace completed requests dynamically to avoid idle slots and maintain high throughput.
 \begin{table}[h]
\centering
\caption{P2P-Agent achieves \textbf{2.1}$\times$ higher token throughput than Ray Data baseline.}
\begin{tabular}{l r r}
\toprule
\textbf{Metric} & \textbf{Ray Data Baseline} & \textbf{P2P-Agent} \\
\midrule
Runtime & 12:57:28 & 17:57:55 \\
Concurrent tasks & 14{,}000 & 14{,}000 \\
Webdoc processed & 9.3M & 25M \\
Questions generated & 410{,}755 & 1{,}192{,}799  \\
Tokens generated & 129{,}622{,}944 & 378{,}591{,}258 \\
Tokens per second & 2{,}778 & 5{,}853 \\
\bottomrule
\end{tabular}
\label{tab:nr_1m}
\end{table}

In Ray Data, decreasing the batch size can partially mitigate idle time. However, each concurrent batch requires a dedicated actor and CPU allocation. Maintaining the same level of task concurrency at smaller batch sizes therefore demands higher data parallelism, which introduces substantial CPU overhead. Moreover, batch-level scheduling incurs additional costs for batch creation and actor management, further compounding inefficiency. Overall, these results demonstrate that fine-grained, row-level scheduling enables more efficient scaling for multi-agent, dynamically controlled workflows than batch-level scheduling in traditional distributed data processing engines.

 % of through asynchronous. Furthermore, the control divergence and the existence of long running tasks within a batch result in inconsistent completion times. A few tasks in a batch may prevent later batches to be scheduled causing resource under-utilization. Matrix’s row-level scheduling approach avoids these inefficiencies similar to the introduction of continuous batch in modern LLM inference engines such as vLLM.
 
\subsection{Tau2-bench}

Tau2-bench~\cite{barres2025tau2} is a recently introduced benchmark for evaluating conversational agents in dual-control environments, where both an AI agent and a user simulator interact with a shared environment through tools and APIs. 
In this experiment, we use Tau2-bench to generate task-solving trajectories for real-world customer support or troubleshooting in the telecom domain. Following prior work such as Kimi K2~\cite{kimiteam2025kimik2openagentic} and AgentBank~\cite{song2024agentbankgeneralizedllmagents}, these trajectories—after filtering and reward validation—can serve as post-training data to enhance LLM reasoning and tool-use performance.

\begin{figure}[h]
    \centering 
    \includegraphics[width=0.7\textwidth]{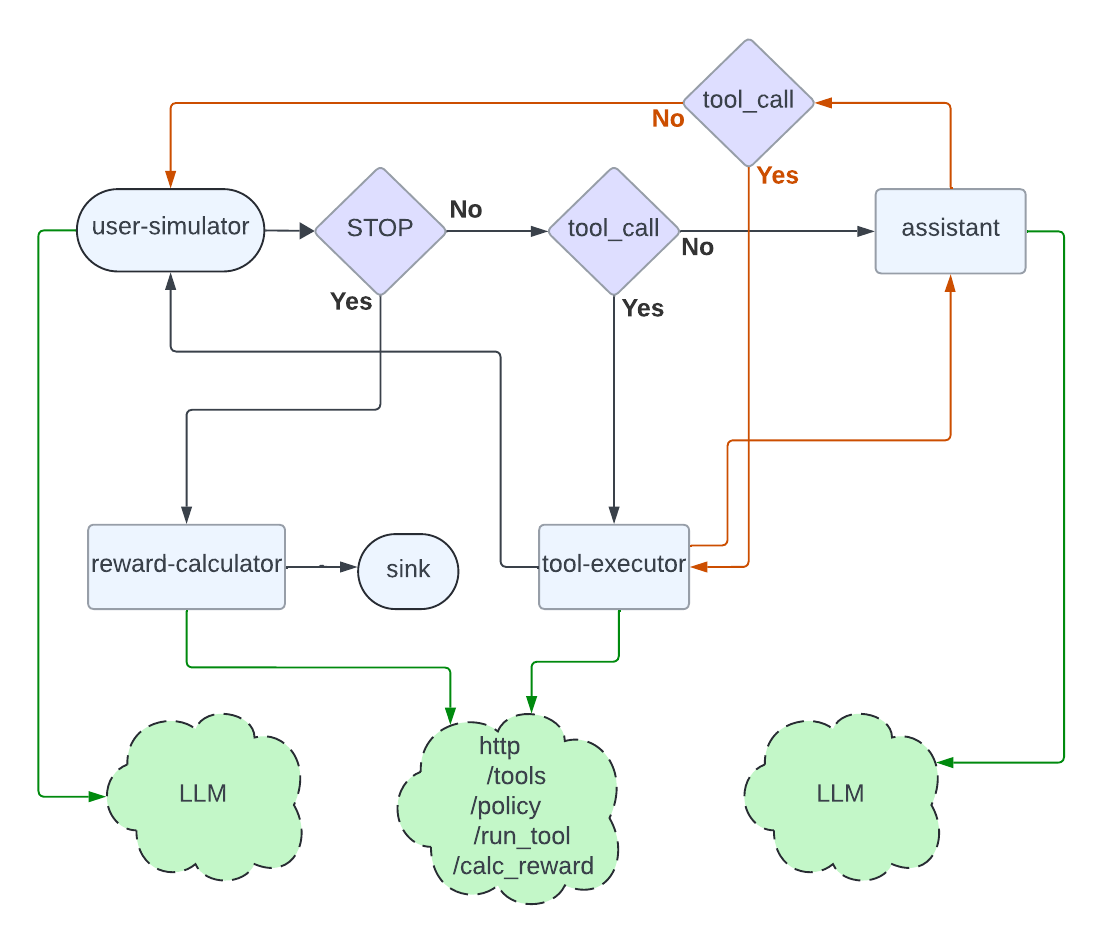}
    \caption{P2P-agent for Tau2-Bench.}
    \label{fig:tau2}
    \vspace{-2mm}
\end{figure}

\paragraph{P2P-Agent Implementation.}
Matrix implements Tau2-Bench as a distributed P2P-agent workflow comprising four functional agents and one orchestrator (Figure~\ref{fig:tau2}).
\begin{itemize}
\item \textbf{User-simulator:} Represents the human user, initiating and responding to the \texttt{tau2-agent}'s queries.
\item \textbf{Assistant:} Acts as the assistant agent, performing reasoning and tool-use steps. % across multiple domains (airline, retail, telecom).
\item \textbf{Tool-executor:} Executes HTTP-based tool calls issued by either the user or assistant. Tool APIs are adapted from the official Tau2-agent implementation~\cite{barres2025tau2_github} and deployed in distributed containers to enable concurrent execution and isolation.
\item \textbf{Reward-calculator:} Validates each trajectory by replaying all tool calls from the initial state and computing task-specific rewards using assertions over the database state. The calculator container reuses the official Tau2-agent implementation, ensuring comparability with benchmark metrics.
\end{itemize}

Matrix exposes two categories of services:
(1) LLM inference services using gpt-oss-120b~\cite{openai2025gptoss120bgptoss20bmodel}, which provide scalable access to model reasoning and dialogue generation, and
(2) containerized task services, derived from Tau2-Bench’s reference implementation.
Each container exposes standardized HTTP endpoints for retrieving tool signatures, executing actions, and evaluating rewards. Service calls are depicted in green in Figure~\ref{fig:tau2}.

\paragraph{Comparison with Tau2 Baseline.}
To evaluate scalability, we compare Matrix’s P2P-agent execution with the official Tau2-agent implementation~\cite{barres2025tau2_github}. The baseline runs all tools and environment logic directly in Python threads on a single node with distributed LLM service. In contrast, P2P-agent distributes agents, LLM and tool-call container services across the Ray cluster. 

As shown in Figure~\ref{fig:trajectory_tau2}, throughput for the Tau2-agent baseline saturates at around 500 threads due to the single-machine constraint. In contrast, P2P-agent continues to scale with concurrency, leveraging distributed placement of agents and containers across the cluster. %, despite the overhead of calling containers.

\begin{minipage}[b]{0.48\textwidth}
    \centering
    \includegraphics[width=0.9\linewidth]{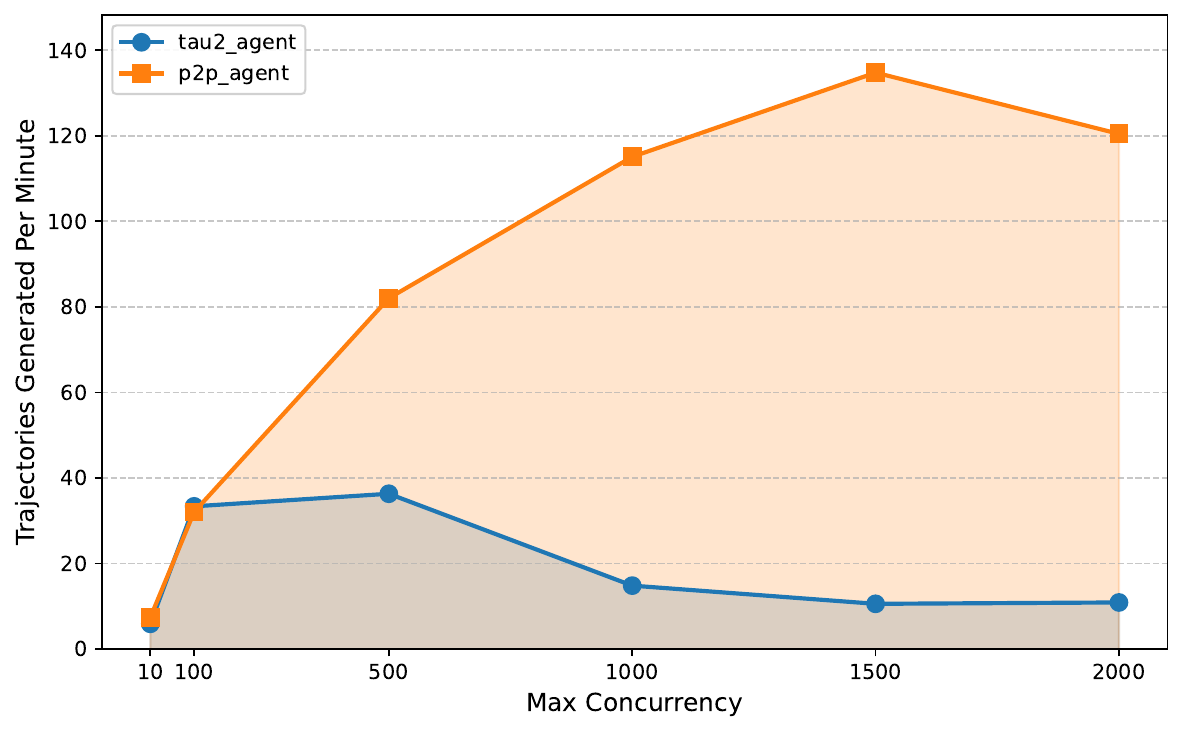}
    \captionof{figure}{Scalability of P2P-agent vs Tau2-agent baseline.}
    \label{fig:trajectory_tau2}
\end{minipage}
\hfill
\begin{minipage}[b]{0.48\textwidth}
    \centering
    \begin{tabular}{l r r}
    \toprule
    \textbf{Metric} & \textbf{Baseline} & \textbf{P2P-Agent} \\
    \midrule
    Runtime & 1:13:41 & 1:15:21 \\
    Concurrent tasks & 500 & 1{,}500 \\
    Total trajectories & 1519 & 22{},800 \\
    Average reward & 0.5918 & 0.5921 \\
    Tokens generated & 11{,}080{,}385 & 185{,}376{,}127 \\
    Tokens per second & 2{,}654 & 41{,}003 \\
    \bottomrule
    \end{tabular}
    \captionof{table}{P2P-Agent achieves \textbf{15.4}$\times$ higher token throughput than Tau2-Agent baseline.}
    \label{tab:tau2_large}
\end{minipage}

\paragraph{Large-Scale Results.}
We further test on 13 H100 nodes, deploying 1.5k containers and 56 gpt-oss-120b replicas. As shown in Table~\ref{tab:tau2_large}, P2P-agent generates $15.4\times$ more tokens per second than the Tau2-agent baseline, while maintaining comparable task rewards. 

\subsubsection{Effect of Message Offloading}
\label{sec:tau2_size}

\begin{figure*}[h]
     \centering
     \begin{minipage}[b]{0.48\textwidth}
         \centering
         \includegraphics[width=0.9\textwidth]{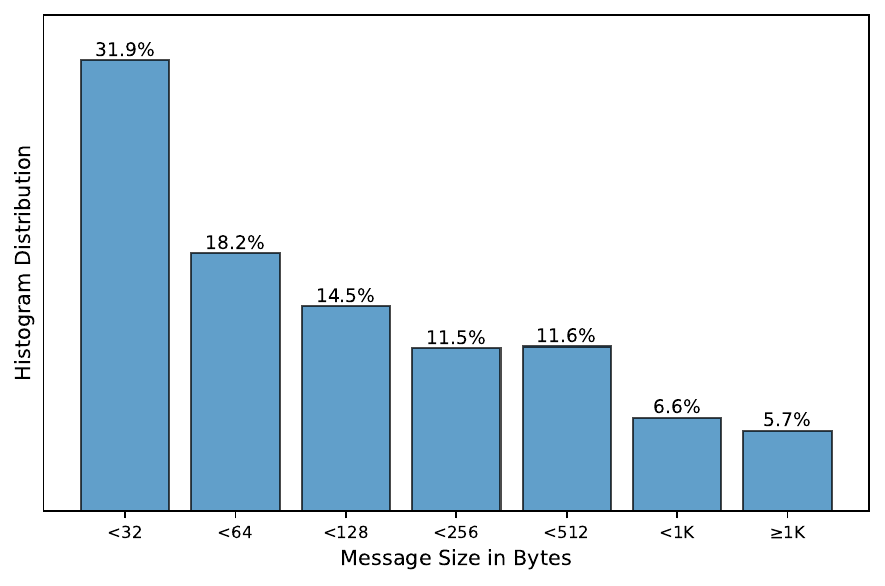}
    \caption{Distribution of conversation sizes in Tau2-Bench.}
    \label{fig:message_sizes}
     \end{minipage}
     \hfill
     \begin{minipage}[b]{0.48\textwidth}
         \centering
    \includegraphics[width=0.9\textwidth]{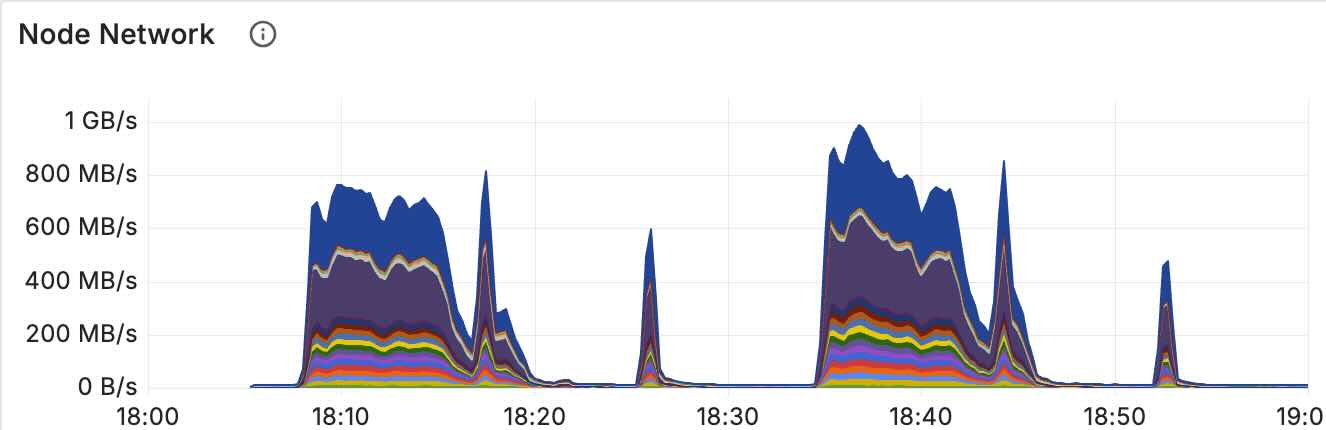}
    \caption{Compare Total Node Network with and without Message Offloading.}
    \label{fig:message_offloading}
     \end{minipage}
\end{figure*}

Matrix orchestrator contains the conversation history. Conversations exchanged in P2P-agent Tau2-bench trajectories vary widely in size, as shown in Figure~\ref{fig:message_sizes}. When orchestrators are routed through distributed agents, large conversation content can cause network overhead and congestion within the cluster.
To mitigate this overhead, Matrix offloads large conversation content to the Ray Object Store, as discussed in Section~\ref{sec:message_offloading}. In this case, contents exceeding 512~bytes are stored in Ray object store and retrieved on demand, which corresponds to about 12\% of the conversations.

Figure~\ref{fig:message_offloading} compares the total cluster network bandwidth during two identical runs: one with message offloading enabled (before 18:30) and one without it (after 18:30). Excluding transient spikes, peak utilization drops from roughly 1~GB/s to 760~MB/s, a reduction of about 20\%. This demonstrates that offloading large conversation contents effectively reduces network traffic and improves scalability under communication-heavy workloads such as Tau2-bench. It also makes the system well suited for future multi-modal data generation tasks.

\section{Conclusion}

We introduced \textbf{Matrix}, a peer-to-peer multi-agent framework for large-scale synthetic data generation. By representing control and data flow as peer-to-peer messages and delegating computation to distributed services, Matrix eliminates centralized bottlenecks and enables efficient execution of tens of thousands of concurrent agent workflows.
Matrix is modular and configurable, allowing users to easily adapt it to diverse data generation tasks and agent roles without modifying core logic. We open-source Matrix at \url{https://github.com/facebookresearch/matrix} to support reproducibility and further research.

\paragraph{Limitations and future work.} Matrix is not a universal fit for all multi-agent workloads. It assumes per-task orchestrator state is serializable and can be passed between agents. It is also less suitable when each step must read/write very large mutable shared state, where data movement or synchronization can dominate costs. Finally, Matrix depends on the Ray actor runtime and therefore inherits its operational constraints and failure semantics.
Looking forward, we plan to provide a library of reusable orchestrator patterns and end-to-end examples, so users can instantiate common control-flow templates with minimal custom code. Future extensions will also explore multi-modal data generation and on-policy continuous data synthesis.

\clearpage
\newpage
\bibliographystyle{assets/plainnat}
\bibliography{agent_paper}

% \clearpage
% \newpage
% \beginappendix

% \section{First appendix}

\end{document}